# Nonnegative Matrix Factorization for identification of unknown number of sources emitting delayed signals

Filip L. Iliev[1], Valentin G. Stanev[1], Velimir V. Vesselinov[2], Boian S. Alexandrov[1,3]*

**1** Theoretical Division, Los Alamos National Laboratory, Los Alamos, NM, United States of America, **2** Computational Earth Science Group, Los Alamos National Laboratory, Los Alamos, NM, United States of America, **3** University of New Mexico Comprehensive Cancer Center, Albuquerque, NM 87102, United States of America

* boian@lanl.gov

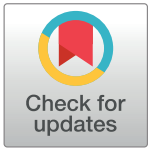

**Data Availability Statement:** An open source Julia language implementation of ShiftNMFk method, that can be used for identification of a relatively small number of signals, and the datasets explored in this paper can be found at: https://github.com/rNMF/ShiftNMFk.jl.

**Funding:** This research was funded by the Environmental Programs Directorate of the Los Alamos National Laboratory. In addition, VVV was supported by the DiaMonD project (An Integrated Multifaceted Approach to Mathematics at the Interfaces of Data, Models, and Decisions, U.S.

## Abstract

Factor analysis is broadly used as a powerful unsupervised machine learning tool for reconstruction of hidden features in recorded mixtures of signals. In the case of a linear approximation, the mixtures can be decomposed by a variety of model-free Blind Source Separation (BSS) algorithms. Most of the available BSS algorithms consider an instantaneous mixing of signals, while the case when the mixtures are linear combinations of signals with delays is less explored. Especially difficult is the case when the number of sources of the signals with delays is unknown and has to be determined from the data as well. To address this problem, in this paper, we present a new method based on Nonnegative Matrix Factorization (NMF) that is capable of identifying: (a) the unknown number of the sources, (b) the delays and speed of propagation of the signals, and (c) the locations of the sources. Our method can be used to decompose records of mixtures of signals with delays emitted by an unknown number of sources in a nondispersive medium, based only on recorded data. This is the case, for example, when electromagnetic signals from multiple antennas are received asynchronously; or mixtures of acoustic or seismic signals recorded by sensors located at different positions; or when a shift in frequency is induced by the Doppler effect. By applying our method to synthetic datasets, we demonstrate its ability to identify the unknown number of sources as well as the waveforms, the delays, and the strengths of the signals. Using Bayesian analysis, we also evaluate estimation uncertainties and identify the region of likelihood where the positions of the sources can be found.

## 1 Introduction

Presently, large datasets are collected at various scales, from laboratory to planetary, [1, 2]. For example, large amounts of data are gathered by distributed sets of asynchronous sensor arrays that, via remote sensing, can measure a considerable number of physical parameters. Usually, each record of a sensor in such array represents a mixture of signals originating from an Department of Energy Office of Science, Grant #11145687). VVV and BSA were supported by LANL LDRD grant 20180060.

**Competing interests:** The authors have declared that no competing interests exist.

unknown number of physical sources with varying locations, speed, and strengths, which are typically also unknown. The analysis of such types of data, especially related to threat reduction, nuclear nonproliferation, and environmental safety, could be critical for emergency responses. One of the main goals of the analyses of such data is to identify and determine the locations of the unknown number of activities, for example by investigating signals of approximately non-dispersive waves, such as: seismic waves, electromagnetic waves in unbounded free space, sound waves in air, radio waves with frequencies less than 15 GHz in air, etc. One way to perform such analysis is to leverage some complex, poorly constrained and uncertain physics-based inverse-model methods where, typically, computationally-intensive numerical models are needed to simulate the governing process related to signal propagation through the medium of interest. An alternative approach, which we will consider here, is the model-free analysis based on the Blind Source Separation (BSS) techniques [3].

In general, there are two widely-used BSS approaches: Independent Component Analysis (ICA), [4, 5], and Nonnegative Matrix Factorization (NMF), [6, 7]. The main idea behind ICA is that, while the probability distribution of a linear mixture of the sources in the observed data is expected to be close to a Gaussian (according to the Central Limit Theorem), the probability distribution of the original sources could be non-Gaussian. Hence, ICA maximizes the non-Gaussian characteristics of the estimated sources, to find the maximum number of statistically independent original sources that, when mixed, reproduce the observation data (see Eq 1). The second approach, NMF, is a well-known unsupervised learning method, created for parts-based representation [8] in the field of image recognition [6, 7], and successfully leveraged for Blind Source Separation, that is, for decomposition of mixtures formed by various types of non-negative signals [9]. In contrast to ICA, NMF does not impose statistical independence or any other constraint on statistical properties (i.e., NMF allows the original sources to be partially correlated); instead, NMF enforces a nonnegativity constraint on the original signals and their mixing ratios. The nonnegativity constraints lead to strictly additive and naturally sparse components that are parts of the data and correspond to readily understandable features. NMF can successfully decompose large sets of nonnegative observations by leveraging the multiplicative update algorithm introduced by Lee and Seung [7], and we use here a modification of this algorithm. However, NMF requires *a priori* knowledge of the number of the original sources.

Another limitation of the classical NMF is that in many real-world problems, especially these concerning physical and environmental processes, there are potential delays of the signals in relation to their arrival at each sensor, leading to a ªshiftº in the recorded mixtures. In another example, a shift in the onset of frequency profile can be naturally induced by the Doppler effect. Therefore, for analyzing: astronomical data, Electroencephalography (EEG) data, Positron Emission Tomography (PET), or fluorescence spectra, taking into account the presence of shifts is beneficial or even necessary. Mathematically, the ªshiftsº means that the signals along the columns of the observation matrix are manifested along different row ranges. A natural extension of NMF is to take into account the potential delays of the signals at the sensors, caused by the different positions of the physical sources in space, combined with the finite speed of propagation of the signals into the considered medium. Various factorization methods have been developed to deal with signals with possible delays or spectral shifts (see, for example, Refs. [10±12]); however, these methods do not consider the situation where the number of the sources producing the delayed signals is unknown. Thus, how to find this unknown number of sources, based only on recorded data with potential delays, remains a largely untreated problem.

Identifying the positions of the sources producing the delayed signals is another common problem for the BSS methods. It arises in many applications (e.g., radars [13], acoustics [14], underwater systems [15]), and it is a natural next step after the determination of the number of sources. Various methods and techniques for addressing this problem have been developed over time, but it is still an area of active research.

Here, we report a new algorithm, called ShiftNMFk, and demonstrate that it is capable of determining the unknown number of the sources of delayed signals and estimating their delays, locations, and speeds based only on records of their mixtures. The main improvement and benefits of our new method is its ability to estimate the unknown number of sources with delays and their locations.

## 2 Methods

### 2.1 NMF minimization for signals with delays

Mathematically, the NMF problem is represented by Eq 1, where the observation data, $V$ ($N \times M$ matrix), is (in the first approximation) a linear mixture of $K$ unknown original signals, represented by $H$, ($K \times M$ matrix), blended by another—also unknown—mixing matrix, $W$, ($N \times K$ matrix), i.e.,

$$V_{n,m} = \sum_{i=1}^{K} W_{n,i} H_{i,m} + \epsilon_{n,m}, \tag{1}$$

where $\epsilon$ ($N \times M$ matrix) denotes the presence of possible noise or unbiased error in the measurements (also unknown). Here, $n$, $i$ and $m$ index the sensors, the sources and the observation moments. These indexes run from 1 to $N$, from 1 to $K$ and from 1 to $M$ respectively, where $N$ is the number of the sensors, $K$ is the total number of unknown sources emitting signals which form the mixtures in the observation data, and $M$ is the number of discretized moments in time at which these mixtures are recorded by the sensors. If the problem is solved in a temporally discretized framework, the goal of the BSS algorithm is to retrieve the $K$ original signals that have produced $N$ mixtures recorded by the given set of sensors. The number of sensors has to be greater than the number of sources.

The observations at different sensors are assumed to be at the same times, however, the temporal spacings between them need not be uniform. If the data are not collected at the same times, interpolation techniques (like cubic splines, see e.g., [16]) can be applied to pre-process the data.

Since both factors $H$ and $W$ are unknown (often even their sizes are unknown because no information about how many constituent signals originating from different sources have been mixed in sensors' records is available), the main difficulty in solving any BSS problem is that it is under-determined.

For NMF to work, the problem must be amenable to a nonnegativity constraint on the sources $H$ and mixing matrix $W$. This constraint leads to the reconstruction of the observations (the rows of matrix $V$) as linear combinations of the elements of $H$ and $W$ that cannot cancel mutually.

The classic NMF algorithm starts with a random initial guess for $H$ and $W$, and proceeds by minimizing the cost (objective) function, $O$, which in our case is the Frobenius norm (it can be also the Kullback-Leibler (KL) divergence, or another feasible norm),

$$O = \frac{1}{2} \| V - W * H \|_F^2 = \frac{1}{2} \sum_{n,m} (V_{n,m} - \sum_{i=1}^{K} W_{n,i} H_{i,m})^2 \tag{2}$$

where, $W_s \geqslant 0; H_{s,n,t} \geqslant 0;$

during each iteration. Minimizing Frobenius norm is equivalent to representing the discrepancies between the observations, $V$, and the reconstruction, $W * H$, as white noise. In order to minimize $O$, it is common to use the gradient descent approach based on the multiplicative updates proposed by Lee and Seung [7]. During each iteration, the algorithm first minimizes $O$ by holding $W$ constant and updating $H$, and then holds $H$ constant while updating $W$. The norm, (Eq 2), is non-increasing under these update rules and invariant when an accurate reconstruction of $V$ is achieved [7].

A limitation of the classic NMF algorithm described above is the assumption of the instantaneous mixing of the signals, which is equivalent to postulating an infinite speed of propagation of the signals in the medium. A natural extension of NMF is to take into account the potential differences between the moments the same signal reaches different sensors [17], caused by the spatial distribution of sensors, combined with the finite speed of propagation of the signals in the medium. One way of treating such type of problems is the approach of a NMF algorithm with shifts [12], designed specifically for signals with delays. Below we describe the key features of this algorithm.

Mathematically, the ªshiftº means that the signals along the columns of $V$ are manifested along different row ranges. The matrix representation of the NMF algorithm for signals with delays includes an additional matrix $\tau$ that incorporates the potential delays (shifts) in observing the same signal at different sensors. A $\tau_{n,i}$ matrix element denotes the delay from the $i^{th}$ original signal to the $n^{th}$ sensor, and using it we can write

$$V_{n,m} = \sum_{i=1}^{K} W_{n,i} H_{i,m-\tau_{n,i}} + \epsilon_{n,m}, \tag{3}$$

In the presence of delays, it is advantageous to use the Fourier space, in which the previous equation becomes

$$\widetilde{V}_{n,m} = \sum_{i=1}^{K} W_{n,i} \widetilde{H}_{i,f} e^{-i2\pi\frac{f-1}{M}\tau_{n,i}} + \widetilde{\epsilon}_f, \tag{4}$$

where the $\widetilde{\cdot}$ denotes the image in Fourier space. Somewhat more succinct version of this equation is

$$\widetilde{V}_f = \widetilde{W}^{(f)} \widetilde{H}_f + \widetilde{\epsilon}_f; \quad \text{where} \quad \widetilde{W}^{(f)} = W_{n,i} e^{-i2\pi\frac{f-1}{M}\tau_{n,i}}, \tag{5}$$

and note that we have used the elementary property of the Discrete Fourier Transform (DFT) to convert time shifts into phase factors. An algorithm that uses the classic NMF strategy of multiplicative updates, but switches back and forth between the time and the frequency domains at each update, has been proposed to find simultaneously $W$, $H$ and $\tau$ [12]. In Fourier space the nonlinear shift mapping becomes a family of DFT transformed $H$ matrices with the shift amount taken by the unknown $\tau$ matrix. Thus the delayed version of the source signal, to the $n^{th}$ channel, is,

$$\widetilde{H}^{(n)}_{i,f} = \widetilde{H}_{i,f} e^{-i2\pi\frac{f-1}{M}\tau_{n,i}}, \tag{6}$$

and the Frobenius norm that has to be minimized is,

$$O = \frac{1}{2}\sum_{n,m}(V_{n,m} - \sum_{i=1}^{K} W_{n,i} H_{i,m-\tau_{n,i}})^2 = \frac{1}{2M} \parallel \widetilde{V}_f - \widetilde{W}^{(f)} \widetilde{H}_f \parallel_F^2, \tag{7}$$

where the last equality holds because of the Parseval's identity. Details of the algorithm of minimization can be found in Ref. [12].

It is important to note that the NMF algorithm with delays determines only the relative delays (the time shifts of the same signal arriving at different sensors). Thus, each $\tau_{i,j}$ is determined up to a constant—a global time shift of all delays would only lead to rearrangement of the matrices in Eq 3. This ambiguity can be simply resolved by centering the delays at zero and splitting them in positive and negative values, which we use below.

### 2.2 Custom clustering for determination of the number of sources

While powerful and producing results easy to interpret the classic NMF method described in the previous section require *a priory* knowledge of the number of the original sources (which we denote by $K$). To address the case when this number is unknown, we have developed an algorithm designed to estimate the number of sources based on the robustness of the minimization solutions. Our algorithm explores consecutively all possible numbers of sources producing signals with delays, from 1, 2. . . to $N$ ($N$ is the number of the sensors) by obtaining a large number of NMF minimization solutions for each number of sources. Then we use clustering to estimate the robustness of each set of solutions corresponding to the same number of sources obtained with different random initial guesses for the minimization. Comparing the quality of the clusters and the accuracy of minimization, we can determine the optimal estimate for $K$. A similar approach has been used for decomposition of instantaneous mixtures of signals (i.e., signals without delays), to analyze the largest available dataset of human cancer genomes [18] and for unmixing hydraulic pressure transients originating from an unknown number of sources [19]. Although our method for estimation of the unknown number of sources is "embarrassingly parallel", for extra-large datasets it will require additional optimizations, such as, distributed arrays and simultaneous execution in parallel of NMF minimization procedure and the custom clustering.

Below we present the clustering method for mixtures of signals with delays. We start by performing $N$ sets of minimizations, which we call NMF runs, one for each possible number $D$ of original sources, which serves to index the distinct NMF models differing only by the number of sources, and goes from 1 to $N$. In each of these runs we have $P$ solutions (e.g., $P$ = 1000) of the minimization with delays for a fixed number of sources, $D$, but with different random initial guesses for the elements of the unknown deconstruction matrices $H$, $W$ and $\tau_D^i$. Thus, each run results in a set of solutions, $U_D$, containing $P$ solutions, each with three matrices, $H_D^i$, $W_D^i$, and $\tau_D^i$,

$$U_D = ([H_D^1; W_D^1; \tau_D^1], [H_D^2; W_D^2; \tau_D^2], ..., [H_D^P; W_D^P; \tau_D^P]), \qquad (8)$$

each of these "3–tuples" represents a distinct solution for the nominally *same* NMF minimization, the difference stemming from to the different initial guesses. Next, we perform a custom clustering to assign each of the $D$ columns of $H_D^i$ ($i$ = 1, 2, . . ., $P$) of these $P$ solutions to one of the $D$ clusters corresponding to one of the $D$ sources. This custom clustering is similar to a $k$-means clustering, but with an additional constraint of holding the number of elements in each of the clusters equal. For example, each one of the $D$ identified clusters (representing the $D$ sources) has to contain exactly $P$ = 1,000 solutions. Note that we have to enforce this condition of having an equal number of points in each cluster since each solution of minimization, specified by a given $[H_D^i; W_D^i; \tau_D^i]$ combination, contributes only one possible solution for each source, and accordingly supplies exactly one element to each cluster.

During the clustering, the similarity between two signals $a$ and $b$, is measured using the cosine distance [20], given by:

$$\rho(a,b) = 1 - \frac{\sum_{i=1}^{n} a_i b_i}{\sqrt{\sum_{i=1}^{n} a_i^2} \sqrt{\sum_{i=1}^{n} b_i^2}}, \quad (9)$$

where $a_i$ and $b_i$ are the individual components of the vectors $a$ and $b$.

After the clustering, we quantify the quality of the clusters obtained for each set by calculating their average Silhouette width [21], and use it to measure how good a particular choice of $D$ is as an estimate for $K$ (the Silhouette widths can vary between −1 and 1). Specifically, the optimal number of sources is picked by selecting the value of $D$ that leads to both (a) an acceptable reconstruction error $R$ of the observation matrix $V$, where $R = \frac{\|V - W*H\|_F}{\|V\|_F}$, and (b) a high average Silhouette width (i.e., an average Silhouette width close to one). The combination of these two criteria is easy to understand intuitively. For solutions with $D$ less than the actual number of sources ($D < K$) we expect the clustering to be good (with an average Silhouette width close to 1), because several of the actual sources could be combined to produce one ªsuper-clusterº, however, the reconstruction error will be high, due to the model being on the under-fitting side (with too few degrees of freedom). In the opposite limit of over-fitting, when $D > K$ ($D$ exceeds the actual number of sources), the average reconstruction error could be quite smallÐeach solution reconstructs the observation matrix very wellÐbut the solutions will not be well-clustered (with an average Silhouette substantially less than 1), since there is no unique way to reconstruct $V$ with more than the actual number of sources, and at least some of the clusters will be artificial, rather than real entities. Thus, our best estimate for the true number of sources $K$ is given by the value of $D = K$ that optimizes both of these metrics. Finally, after choosing the optimal $D = K$, we can use the centroids of the $K$ clusters of $U_K$, that is: the waveforms of the signals of the unknown sources, $H_K$, and the corresponding centroid representing the mixing matrix, $W_K$ and the delay, $\tau_K$, to derive a robust solution of the NMF problem with delays.

## 2.3 Retrieving the locations of the sources

By applying our clustering algorithm we determine the number of the sources, and from the minimization (with the determined number of sources) we obtain the waveforms, the mixing ratios and the delays associated with each of the sources. With this information available, we can estimate the locations of the sources and speed of propagation of the signals. Let's denote by $r_{j,i}$,

$$r_{j,i} = \sqrt{(x_j^s - x_i^d)^2 + (y_j^s - y_i^d)^2}, \quad (10)$$

than the objective function $F$ is defined like this:

$$F = \sum_{j}^{n} \sum_{i \neq i^*}^{K} \left( \frac{(\tau_{j,i^*} - \tau_{j,i}) - \frac{(r_{j,i^*} - r_{j,i})}{v_j}}{\sqrt{\sigma_{j,i^*}^2 + \sigma_{j,i}^2}} \right)^2. \quad (11)$$

To minimize $F$ we use a weighted Least-squared minimization procedure, and estimate the coordinates of the sources based on coordinates of the sensors and the signal delays. In $F$, $i^*$ denotes the closest sensor to the source with number $j$ (the sensor with the smallest delay for $j^{th}$ source), and $K$ and $n$ are the number of sources and sensors, respectively. For simplicity, we assume that the speed of propagation of each of the signals, $v_j$, is the same ($v_j = v$), but, in general, the method is not restricted to this particular case. Also, note that we have assumed a two-dimensional medium, but $F$ in more general form can be used in arbitrary dimensions provided that $r_{j,i}$ is suitably modified. Further, the equation for $r_{i,j}$ gives the distance from a source $j$ to sensor $i$, and $x_j^s$ and $y_j^s$ are the coordinates of source $j$, while $x_i^d$ and $y_i^d$ are the coordinates of sensor $i$. The standard deviations, $\sigma_{j,i}$, are the sample standard deviations of $\tau_{j,i}$ for a fixed source $j$ and sensor $i$ in the run. The ªsamplesº are obtained from each run of the minimization procedure, and we assume that all the $\tau_{j,i}^p$, $p = 1, 2, \ldots, P$ derived by the minimization in the runs are normally distributed.

The coordinates of all the sensors are known. The coordinates of the sources are the unknown variables, which, along with the speeds of signals propagation, can be found by minimizing the objective function $F$. Indeed, the difference between the signal's delays to two (arbitrary) sensors, $i_1$ and $i_2$, should be the same as the difference between the corresponding distance $r_{1,2}$ between these sensors divided by the speed of propagation. Hence, by minimizing $F$, we are trying to find the coordinates of the sources and speeds that make this true. Minimizing $F$ also allows us to propagate the uncertainty/error of the NMF minimization estimates into our source location estimates by incorporating the sample standard deviations of $\tau_{j,i}$ for each source.

### 2.4 Implementation of our algorithm

So far we have outlined the key parts of the proposed algorithm. In this section, we concentrate on providing a detailed description of its implementation. Our algorithm includes the following procedures: **Minimization + Elimination criteria + Custom clustering + Optimization for retrieving the locations of the sources + Uncertainty analysis**.

Below we provide some implementation details concerning these procedures.

**2.4.1 Minimization procedure.** The starting point of our method is the modification of the NMF minimization designed for signals with delays [12].

The mathematical representation of the minimization of delayed signals is similar to the classic NMF multiplicative algorithm but includes an additional matrix $\tau$ that incorporates the potential delays (shifts) in observing the same signal at different sensors. The matrix element $\tau_{n,i}$ denotes the delay from the $i^{th}$ original signal to the $n^{th}$ sensor and using it we can write,

$$V_{n,m} = \sum_i W_{n,i} H_{i,m-\tau_{n,i}} + \epsilon_{n,m}, \tag{12}$$

The minimization simultaneously determines $W$, $H$ and $\tau$ matrices, via switching back and forth between the time and the frequency domains at each update. The updates of the mixing matrix, $W$, is done in the same way as the classic NMF but incorporating one of the changed (by $\tau$) $H^n$ matrix (which is also nonnegative). Because the delays are unconstrained, the shift matrix $\tau$ is estimated by Newton-Raphson method which looks for the minimum of a function with the help of its gradient and Hessian [12].

We explored the optimal number of iterations, $I$, needed to have a reasonable reconstruction error; $R = \frac{\|V - W*H\|_F}{\|V\|_F}$. Our results demonstrate that after a certain number, $I_{max}$, increasing further the number of the iterations does not lead to any improvement in the final results. This is because often the algorithm stops by its internal convergence criteria, before reaching the maximum number of iterations. We found that, $I_{max} = 50,000$, produces acceptable results.

**2.4.2 Impact of pair signals' correlations.** To be able to determine the number of the original sources we need to perform a large number of simulations (to build the clusters) and, hence, we need a proper understanding of the limitations of the minimization with different random initial guesses for the elements of $W$ and $H$ is required. To unravel these limitations,

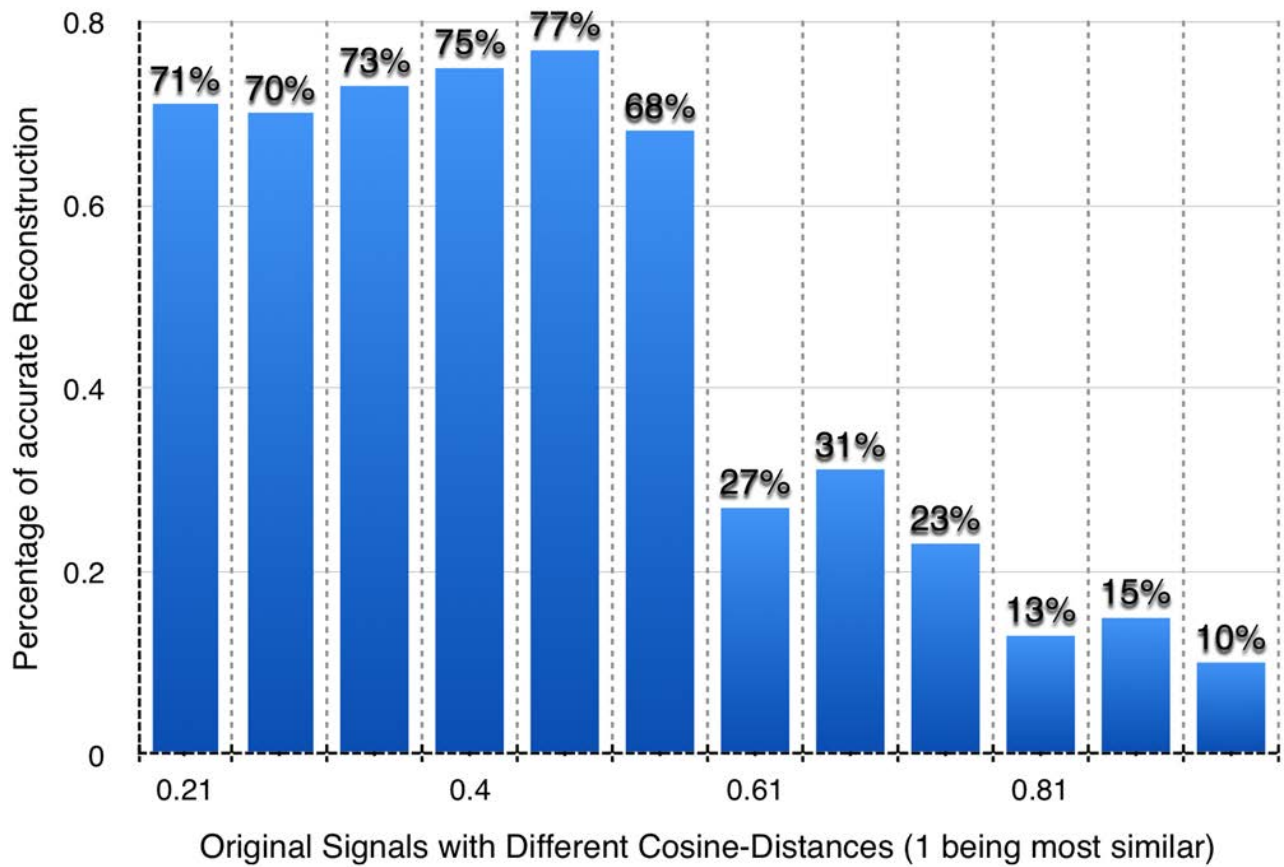

**Fig 1. The connection between the ratio of the number of successful recognitions and similarity of the signals.** The bars on Fig 1 represent the percentage of the correct reconstructions obtained by the minimization, for the case of 3 source signals with different levels of cosine similarity between two of them.

https://doi.org/10.1371/journal.pone.0193974.g001

we performed a large number of minimizations with different random initial guesses for observation data generated with waveforms with different level of pair correlation. Our results demonstrate that minimization works better if the original signals are not very strongly correlated. The reason is easy to understand. If the waveforms of two of the sources are strongly correlated, the algorithm can easily assume that the mixtures recorded by the sensors are produced by one source instead of two nearly identical sources, thus returning a wrong number of original signals that, however, provides a decent reconstruction, $R$.

Changing the correlation between two of three initially generated signals, we studied the success rate of the reconstruction of the minimization by comparing the distance between the generated wave-forms and these derived by the minimization. For this comparison we used the cosine distance. For a successful recognition, we accept cosine distance [20], between two signals, $\rho(a, b) \geq 0.95$.

Fig 1 presents the relationship between the similarity of the original sources (measured by cosine distance) and the success rate (ratio of the number of successful recognitions to the total number of solutions) in groups containing one hundred minimizations, for sources with the same correlations. Our results demonstrate that for relatively uncorrelated source signals the success rate is above 70%. This rate experiences a sharp drop off to around 30% for signals with greater than 0.6 cosine similarity, and further decreases to 10%, when the similarity exceeds 0.8.

**2.4.3 The elimination criteria.** When performed many times with fixed number of sources, $D$, but with random initial guesses, the minimization either converges to different (sometimes very dissimilar) solutions or stops (fails) before reaching a good solution. Considering the ill-posedness of the minimization problem this behavior is expected. The minimization behavior depends on various factors, such as the initial guesses, the ratio between the number of sources and the number of sensors, the specific shape of the waveforms of the signals and delays, etc. For example, achieving a good minimization and obtaining an accurate reconstruction depends on the level of correlation between the waveforms of the original signals (see, Fig 1. As a result, when we performed a large number of consecutive minimizations with different initial conditions, the algorithm many times returns solutions that demonstrate a poor reconstruction of the observation matrix, $V$.

To overcome this problem, we employ two elimination criteria needed to extract from each set of solutions (from a reasonably sized pool of $P$ minimizations) only those that both reproduce accurately the observations and are physically meaningful. Specifically, we use the following two elimination criteria to discard (a) minimization outliers that do not reconstruct sufficiently well the observation matrix, $V$. Further, (b) we discard the minimization solutions that do not satisfy a general physical condition; the physical condition we are imposing is a constraint related to the existence of a maximum signal delay (between all the sensors) that arises from the finite size of the sensor array and finite speed of propagation of the signals in the medium. Let's consider in details these two elimination criteria.

1. *Removing the outliers*: We discard the worst 10% of the solutions of the minimization by ordering them with decreasing discrepancy between the observation matrix, $V$, and its reconstruction, $W * H$. This criterium removes solutions of the minimization that are crude representations of the observation matrix, $V$, and can be considered outliers.

2. *Imposing maximum time delay*: Our second criterion for a solution elimination explores the size and spread of the estimated signal delays. Physically, in a given sensor array, the sensors are separated by various distances. The largest of these distances we will denote by $L_{max}$. We define the maximum time for travel through the array, $t_{max}$, as the time need for a signal to propagate the largest distance between two sensors in the array: $t_{max} = L_{max}/v$. Therefore, for a given signal, the differences between any pair of the estimated signal delays for any pair sensors cannot be larger than the time, $t_{max}$ (the time needed for this signal to propagate the maximum distance, $L_{max}$). Based on that, our second elimination criterium is: the spread in the estimated delays for each of the signals at all the sensors is bounded by the maximum travel time, $t_{max}$.

To implement this second criterion, we need information about the physical coordinates of the sensors in the array (typically, this information is known) and the propagation speed of the signals (that could be unknown). However, an approximation of this criterion can be implemented if the propagation speed is unknown. This approximation is achieved by evaluation of the standard deviation, $std(\tau_{j,i})$, and the average value of the delays, $mean(\tau_{j,i})$, of all the estimates associated with a given signal $j$. In particular, we calculate the coefficient of variation, $CV(\tau_{j,i})$ of the delays from all the sources to all the sensors, $CV(\tau_{j,i}) = std(\tau_{j,i})/mean(\tau_{j,i})$. If the spread of the delays of a given signal over all the sensors (in the simulations in a given run), $CV$, is greater than a certain cut-off (here $CV \geq 0.8$) the corresponding solution is considered unphysical and rejected.

The described two elimination criteria are sufficiently general to be valid for various types of problems. However, details of their implementation and the exact choice of $R$ and $CV$ cut-offs may depend on particular problem-specific details such as the geometry of sensors and sources, noise levels, etc.

**2.4.4 Bayesian analysis of the posterior uncertainties.** To obtain the posterior probability distribution functions (PDFs) of the estimated signal propagation speed and the coordinates of the sources, we use a Bayesian analysis based on Markov Chain Monte Carlo (MCMC) sampling. The PDFs are obtained based on the estimated delays and following Bayes theorem [22, 23], using a likelihood function defined as $e^{(-\chi^2/2)}$. The analysis was performed using the Robust Adaptive Metropolis Markov Chain Monte Carlo (MCMC) algorithm [24], implemented in the open-source computational framework MADS (Model Analysis and Decision Support; http://mads.lanl.gov) and applied as in [25]. Using the results of this Bayesian analysis, we can define a region of likelihood where the positions of the original sources can be found with a given probability.

## 3 Results

Here we demonstrate that the application of our algorithm ShiftNMF*k* to several synthetic datasets indeed leads to successful identifications of the unknown number of sources, their locations, the relative signal strengths, the signal propagation speed. We also estimate the posterior uncertainties of the estimated source locations and the signal propagation speed.

### 3.1 Generation of synthetic datasets for reconstruction of the original sources

To verify the method we constructed several synthetic datasets by generating various observation matrices using a semi-random approach. To generate the initial synthetic datasets we do not need the speed of the signals. Specifically, we start with several (two, three, and four) basic waveforms, as original signals, $H$, which we mix and shift, with respect to each sensor, by randomly generating the mixing matrix, $W$, and delay matrix, $\tau$. We are using three sensor arrays with 16, 18, or 24 sensors distributed on a rectangular discrete lattice with a distance between two sensors along the lattice equal to one. In this way, for each random choice of the sources matrix, $H$, the mixing matrix, $W$, and the delay matrix $\tau$, we obtain a different observation matrix, $V$. We explored the success rate of our algorithm on several synthetic problems generated in this way.

### 3.2 Finding the unknown number of sources and reconstructions of the original signals with delays

**3.2.1 Reconstruction of 3 sources/signals using 18 sensors.** We begin with 3 pre-defined signals mixed and delayed randomly to produce a test case with observations recorded by 18 sensors. Fig 2 shows the resulting robustness of the solutions based on the obtained solution clusters (as discussed in Section 2.2 above). The comparison of the average Silhouette width (in blue) for $D$ ranging from 1 through 6 sources, shows how well our solutions are clustered. We observe that there is *no good clustering* for more than 3 sources. The red curve shows how well the average reconstruction errorÐthe difference between the original observations and the reconstructionÐis minimized. Again, the best result with average reconstruction error closest to zero is achieved for more than 3 sources. Based on this, we conclude that the system has $K = 3$ original sources mixed at 18 sensors. The lower plot shows the derived signals and their mixing matrix and respective delays as compared to the true signals used to generate the example.

**3.2.2 Reconstruction of 4 sources/signals using 24 sensors.** Fig 3 shows the results from simulations with 4 sources and 24 sensors. The comparison of the average Silhouette width (in blue) for 1 through 6 number of sources, shows how well our solutions for the original signals are clustered together. We see the best clustering occurs for $K = 4$ sources, and this is where the average reconstruction error is minimized as well. Thus, we can conclude that the system has 4 original sources mixed at 24 sensors. The lower plot shows the derived

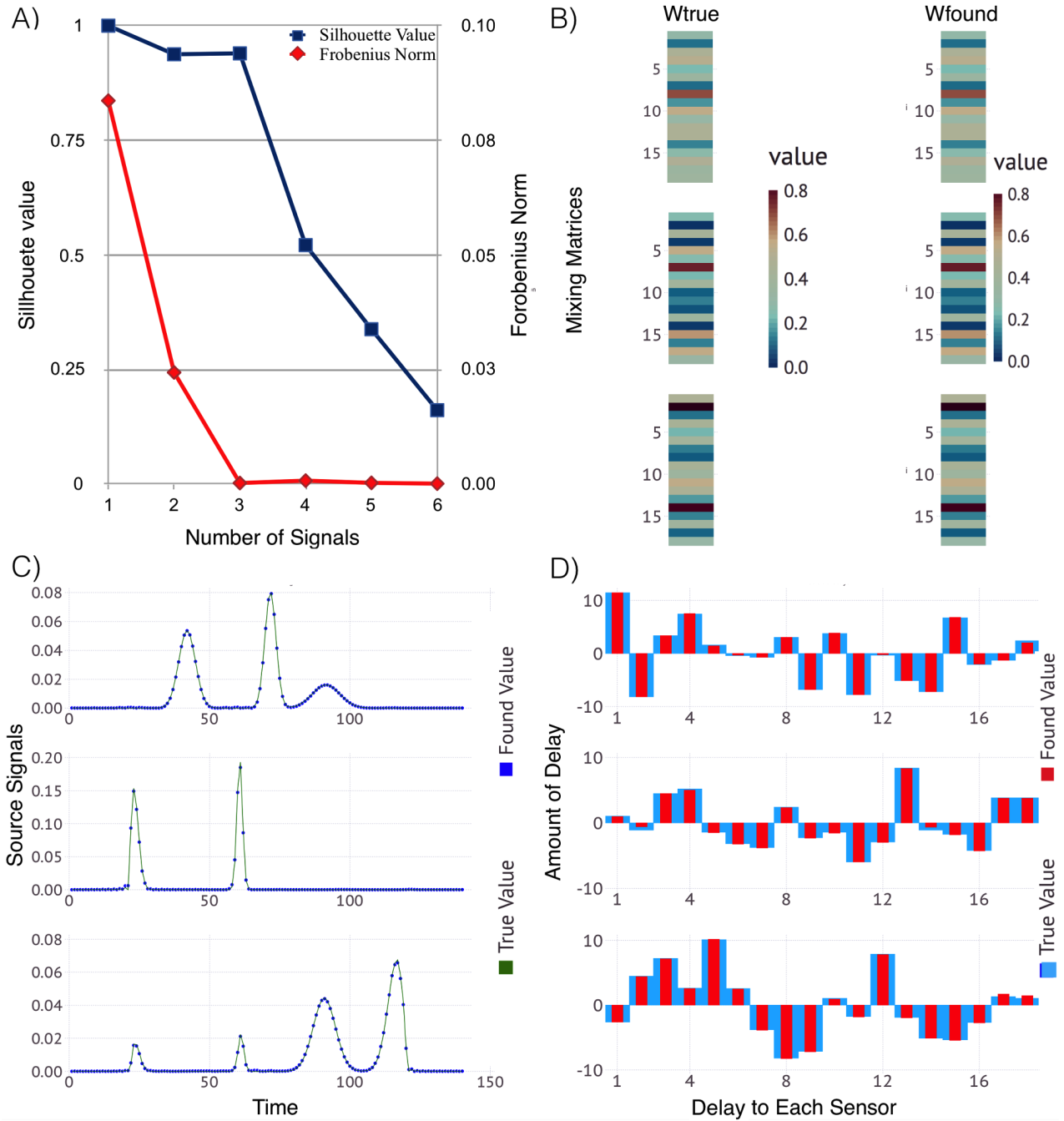

**Fig 2. ShiftNMF*k* results identifying the unknown; number of sources, waveforms of the signals, delays and mixing weights for 3 sources and 18 sensors.** Panel A: The average silhouette width (blue) and the average reconstruction error (red) for a different number of sources. Panel B: Color coded visualization of the mixing matrices (true and estimated) where each column represents how well each source is represented at each sensor. Panel C: The true (green line) and the estimated (blue dots) signal transients superimposed over each other. Panel D: The signal delays at each sensor represented as a bar graph with true (blue bars) and estimated delays (red bars).

https://doi.org/10.1371/journal.pone.0193974.g002

signals and their mixing matrix and respective delays as compared to the true values used to generate the example.

The above analyses demonstrate the applicability of our new protocol for identification of unknown delayed sources based on Shifted Nonnegative Matrix Factorization technique

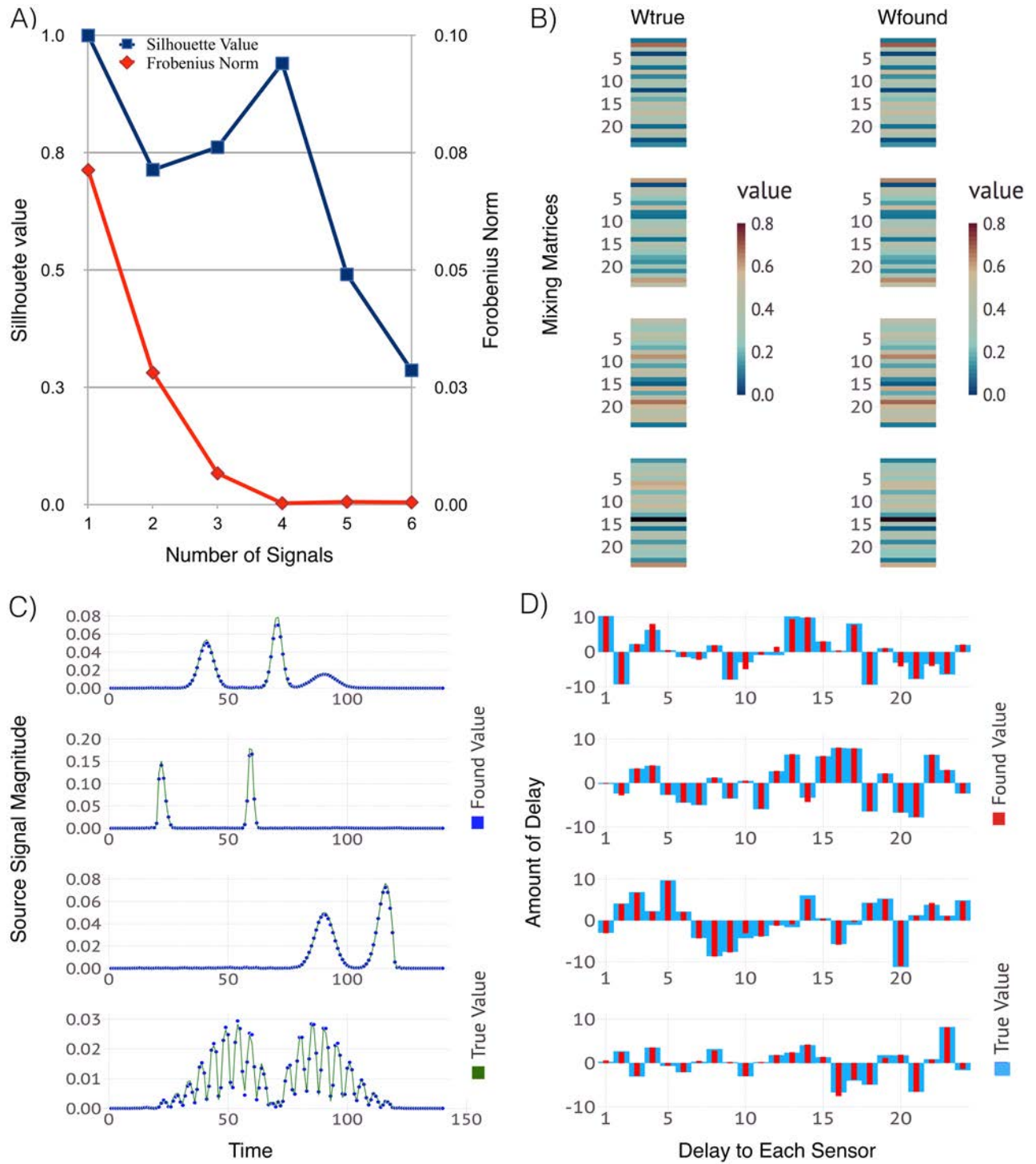

**Fig 3. ShiftNMF*k* results identifying the unknown; number of sources, waveforms of the signals, delays and mixing weights for 4 sources and 24 sensors.** Panel A: The average silhouette width (blue) and the average reconstruction error (red) for a different number of sources. Panel B: Color coded visualization of the mixing matrices (true and estimated) where each column represents how well each source is represented at each sensor. Panel C: The true (green line) and the estimated (blue dots) signal transients superimposed over each other. Panel D: The signal delays at each sensor represented as a bar graph with true (blue bars) and estimated delays (red bars).

https://doi.org/10.1371/journal.pone.0193974.g003

combined with a custom semi-supervised clustering, minimization and elimination procedures. More examples can be explored by using our implementation of ShiftNMF*k* at https://github.com/ShiftNMFk.jl.

### 3.3 Identification of the source locations and the signal propagation speed

To demonstrate the capabilities of our algorithm to identify unknown number of sources as well as their locations and propagation speed, we generate additional synthetic datasets. In this case, we build the observation matrix, *V*, by applying preselected source locations, waveforms and amplitudes, and propagate their signals to the sensors with a preselected speed. Here, instead of generating random mixing matrix, *W*, and delay matrix, *τ*, we calculated them as a function of the source positions, the signal propagation speed and the signal amplitude decays.

Here we present the results for 3 sources arbitrary placed in a lattice with 16 evenly spaced sensors (the side of the lattice was set to 1), for two cases; with the sources inside and sources outside of the sensor lattice. The waveforms of the signals were chosen to be the same as those shown in Fig 2 and with arbitrary amplitudes between 0 and 1 and a specific signal-amplitude-decay function following a $1/\sqrt{r}$ law. This particular type of decay function has relevance to a real-world problem: $(1/\sqrt{r})$ is characteristic of the decay of the amplitude of two-dimensional surface waves in three-dimensional space (e.g., Rayleigh seismic waves). Other possible common forms (not shown here) are $1/r$ and $1/r^2$, reflecting the physics of wave propagation in a three-dimensional space—they describe the decay of the amplitude and the intensity (which goes as the square of the amplitude) of a wave with the distance from the source. Note that in all the simulations the signals keep their original waveforms the same as they propagate in space, i.e. the waves are considered to be non-dispersive.

The results from ShiftNMF*k* simulations are presented in Figs 4 and 5. It can be seen that the estimates are very close to the ones used to generate these two synthetic datasets.

The estimates for the source coordinates for the two cases, obtained by minimization of *F* (Eq 11) (see, Methods), are presented in Fig 6, with uncertainties derived by Bayesian analysis. The minimization of *F* is performed using the NLopt.jl Julia package and running the non-linear minimization ∼1,000 times with different initial (random) guesses for the unknown parameters. After each minimization, we remove the outliers using two steps. First, the worst half of the solutions of minimization is removed. Next, from the remaining solutions, we estimate the median source position coordinates and we remove additional 50% of the worst solutions based on the discrepancy from the median coordinates. In this way, for each source, we result with relatively tightly clustered solutions around the likely coordinates.

The calculations of the posterior uncertainties were performed using the LANL computational framework MADS [26]. In particular, we constructed regions of likelihood (uncertainties) around the estimated coordinates of the sources. The results are presented in Figs 7 and 8. From the figures, It can be seen that the location of a given source and the geometry of the sensor array could lead to a much narrower uncertainty along some direction, which manifests itself in non-zero correlation coefficients between *x* and *y* coordinates for some sources.

Table 1 provides the summary of our results, with standard deviations obtained from the Bayesian Analysis. The propagation speed of the signals has been estimated (by the minimization) to be: $v = 0.500 +/- 0.001$.

## 4 Discussion

We have developed a new method, called ShiftNMF*k*, for blind source separation and identification of the unknown number, locations, strengths and signal propagation speeds for sources emitting non-dispersed signals with delays, which produce signal mixtures that are

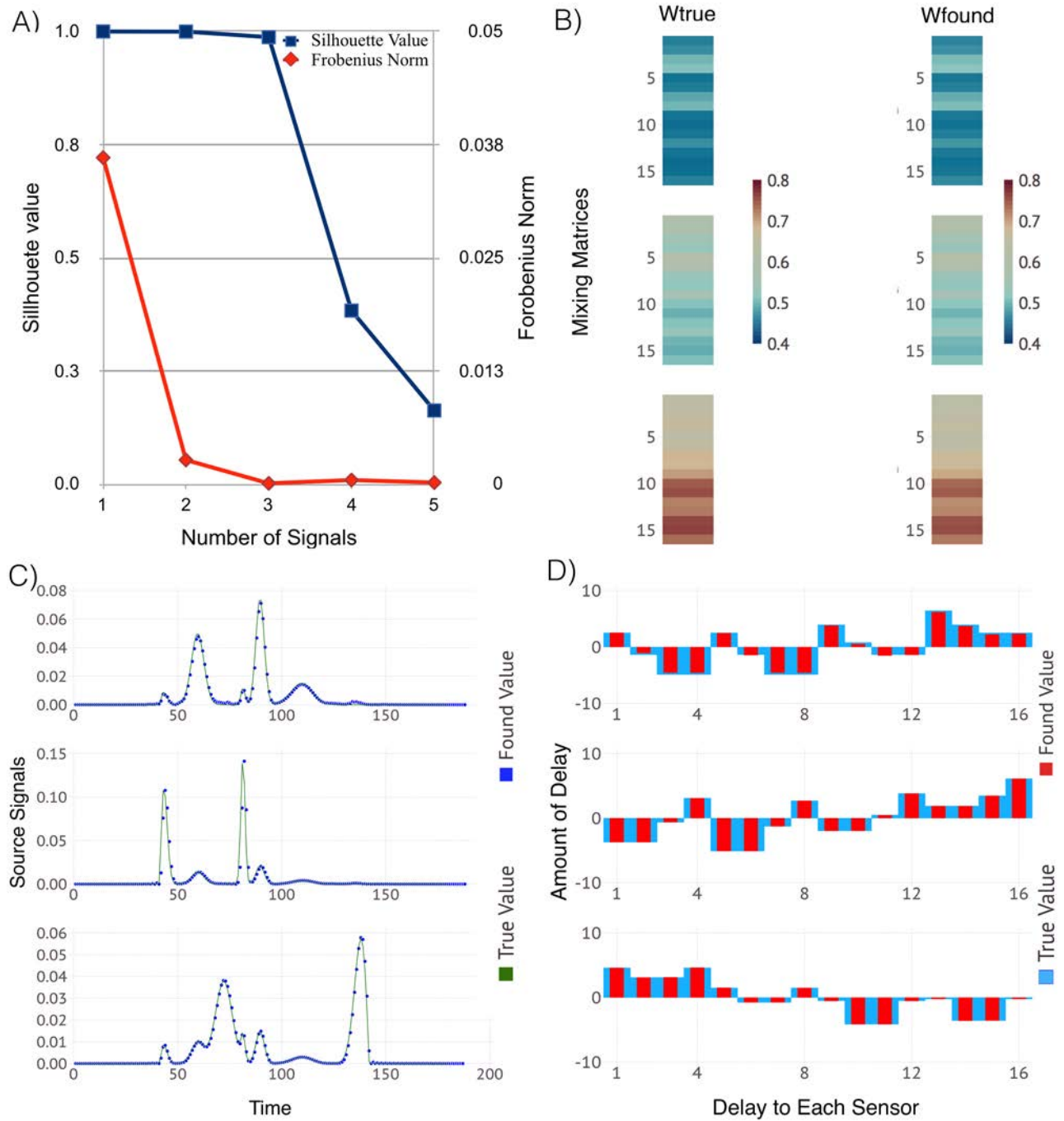

**Fig 4. ShiftNMF*k* results for identification of the locations and speed of unknown number of sources *inside* the sensor array.** Panel A: The average silhouette width (blue) and the average reconstruction error (red) for a different number of sources. Panel B: Color coded visualization of the mixing matrices (true and estimated) where each column represents how well each source is represented at each sensor. Panel C: The true (green line) and the estimated (blue dots) signal transients superimposed over each other. Panel D: The signal delays at each sensor represented as a bar graph with true (blue bars) and estimated delays (red bars). The correct number of sources is identified (3). The estimates for the signal transients, the signal delays and the signal mixing are very close to their true values.

https://doi.org/10.1371/journal.pone.0193974.g004

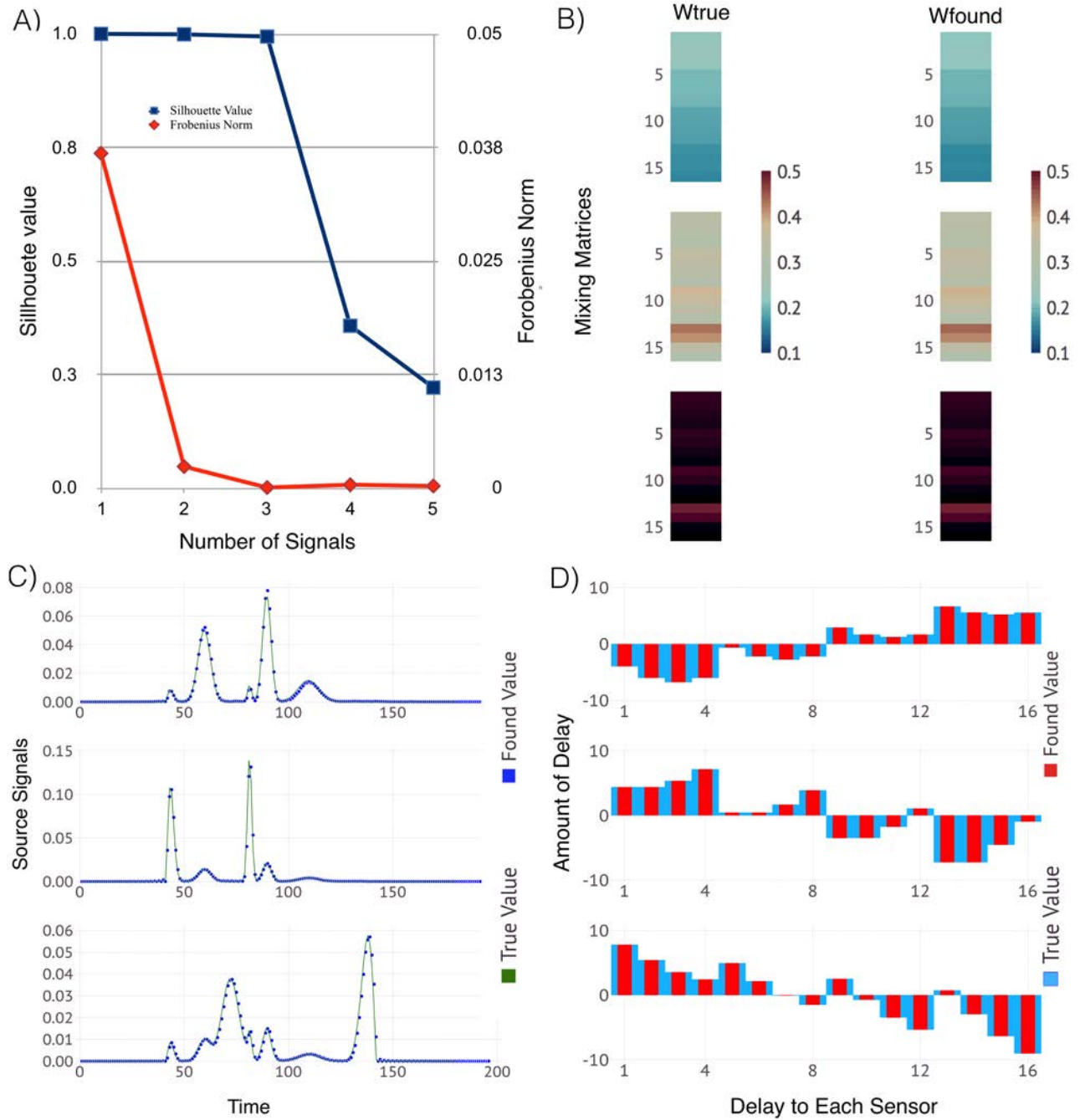

**Fig 5. ShiftNMF*k* results for identification of the locations and speed of unknown number of sources *outside* the sensor array.** Panel A: The average silhouette width (blue) and the average reconstruction error (red) for a different number of sources. Panel B: Color coded visualization of the mixing matrices (true and estimated) where each column represents how well each source is represented at each sensor. Panel C: The true (green line) and the estimated (blue dots) signal transients superimposed over each other. Panel D: The signal delays at each sensor represented as a bar graph with true (blue bars) and estimated delays (red bars). The correct number of sources is identified (3). The estimates for the signal transients, the signal delays and the signal mixing are close to the true values.

https://doi.org/10.1371/journal.pone.0193974.g005

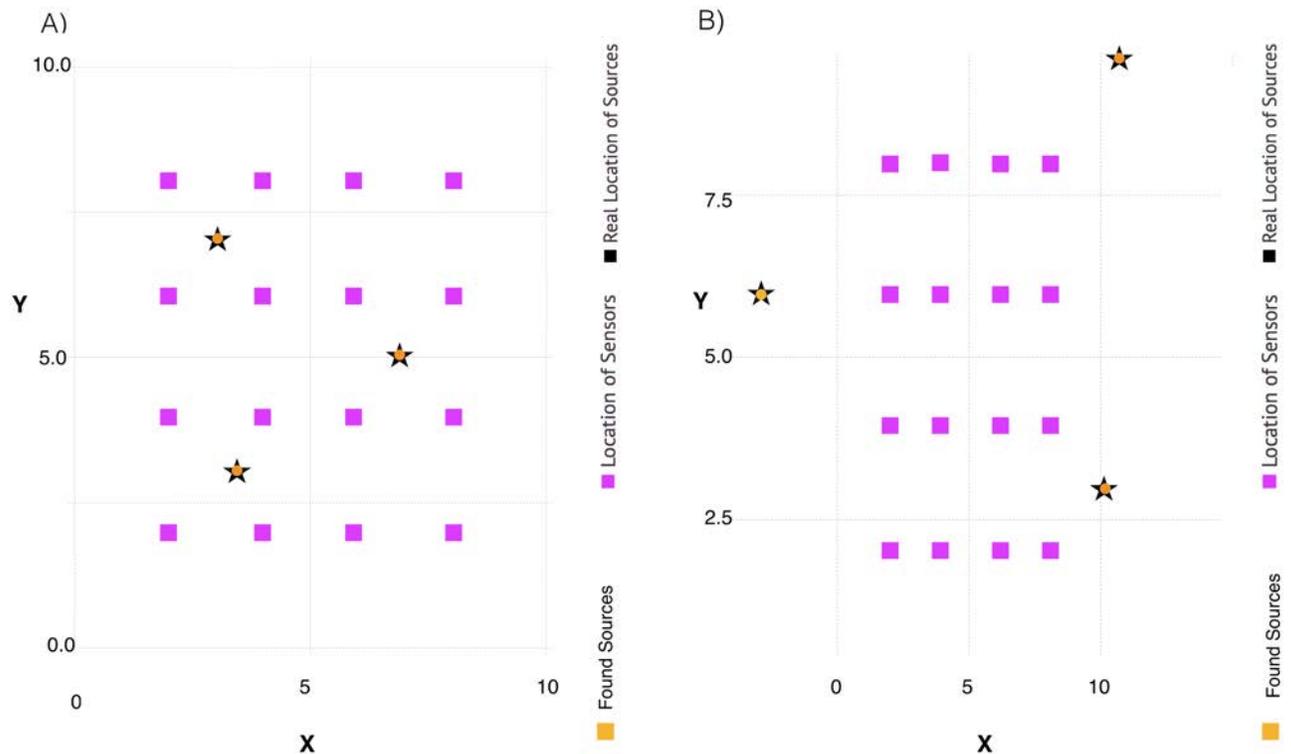

**Fig 6. ShiftNMF*k* results for retrieving the sources' locations.** Panel A: The unknown number of sources are inside the sensor array; Panel B: The unknown number of source are outside the sensor array.

https://doi.org/10.1371/journal.pone.0193974.g006

recorded by a set of sensors. Our work provides a useful tool for blind source separation because (1) the original NMF algorithm does not work with signals with delays and (2) the available NMF algorithms that account for delays require *a priory* knowledge of the number of the signals.

Application of our method to synthetic datasets demonstrate its applicability for identification of an unknown number of delayed sources and their properties, based on an advanced minimization procedure which is combined with (a) two selection criteria for rejecting anomalous and physically implausible solutions, (b) a custom semi-supervised clustering for identification of the unknown number of sources based on the solution robustness, (d) another minimization procedure for identification the source locations, and (d) Bayesian analysis of the posterior uncertainties. ShiftNMF*k* can also identify the signal propagation speeds based on the derived signal delays and information about sensor locations. Additional physical knowledge about the type of the signals and physics of signal propagation (specifically, how the amplitude of the signal changes with the traveled distance) can be leveraged to increase the efficiency and speed-up the convergence of the source-locating procedure.

For the generated synthetic data sets, the unknown number and locations of the sources are identified from a set of mixed signals recorded by arrays of monitoring sensors, without any additional information about (1) the number of sources, (2) source locations, (3) source propagation speed, or (4) source delays.

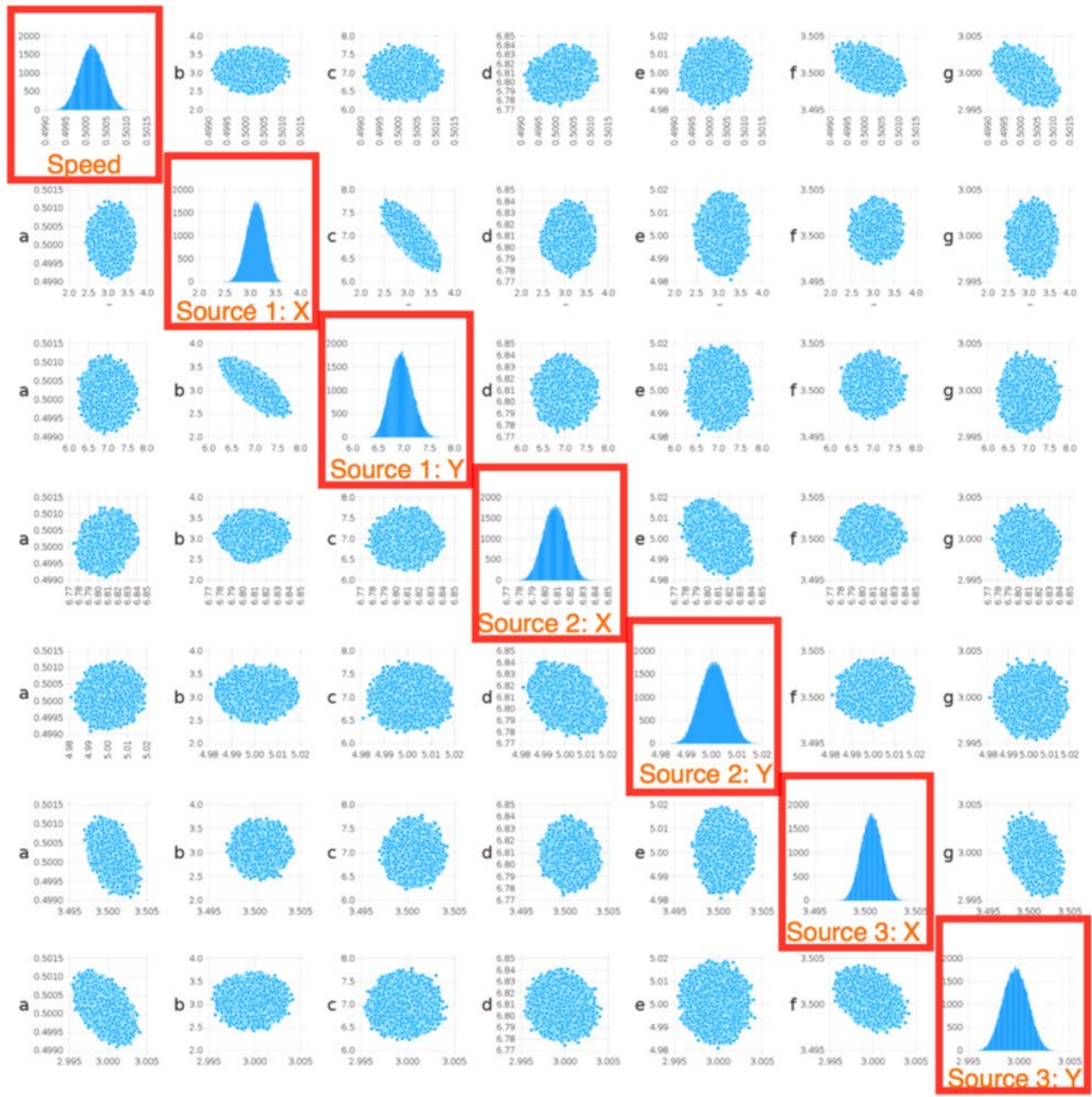

**Fig 7. Bayesian analysis of ShiftNMF*k* estimates for the case of 3 sources *inside* an array of 16 sensors.** The highlighted subplots show the distribution of the variables estimated by the location minimization procedure: the signal propagation speed and the *x* and *y* locations for each of the 3 sources.

https://doi.org/10.1371/journal.pone.0193974.g007

It is important to note that the solved here inverse problem is underdetermined (ill-posed). To address this, the presented algorithm explores the plausible solutions and seeks to narrow the set of possible solutions by robust and parsimonious analysis of the observed data. An open source Julia language [27] implementation of ShiftNMF*k* method, that can be used for identification of a relatively small number of signals, and the datasets explored in this paper can be found at: https://github.com/rNMF/ShiftNMFk.jl.

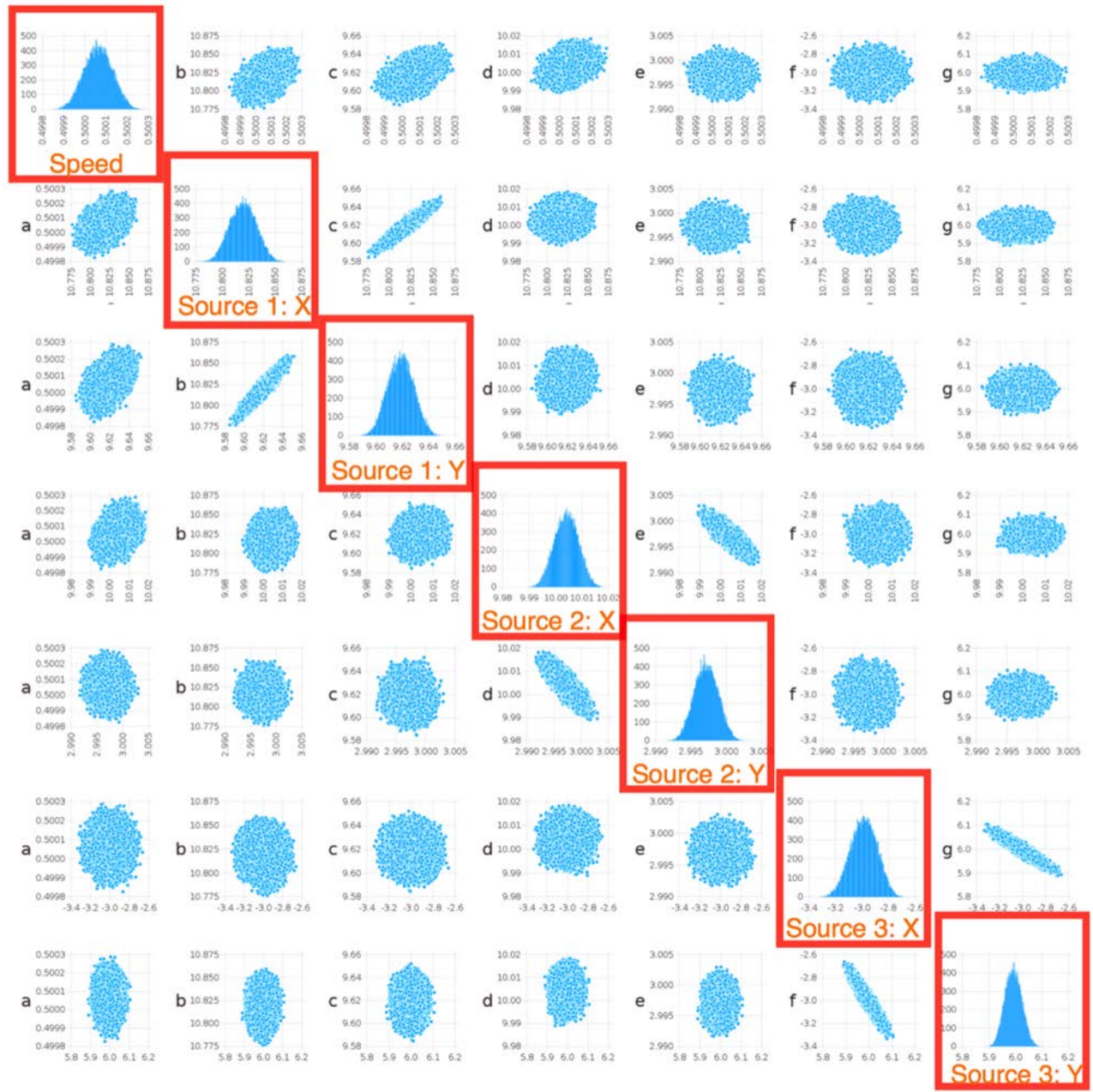

**Fig 8. Bayesian analysis of ShiftNMF$k$ estimates for the case of 3 sources *outside* an array of 16 sensors.** The highlighted subplots show the distribution of the variables estimated by the location minimization procedure: the signal propagation speed and the $x$ and $y$ locations for each of the 3 sources.

https://doi.org/10.1371/journal.pone.0193974.g008

**Table 1. ShiftNMF*k* estimates for the source coordinates for the cases of sources inside and outside the sensor array with the standard deviations obtained by Bayesian analysis.**

| source position | $x_{true}$ | $x_{NMF} \pm 2\sigma_x$ | $y_{true}$ | $y_{NMF} \pm 2\sigma_y$ |
|---|---|---|---|---|
| inside the array | 3 | 3.11 ± 0.07 | 7 | 6.95 ± 0.09 |
|  | 3.5 | 3.50 ± 2.1e-6 | 3 | 2.99 ± 2.9e-6 |
|  | 6.8 | 6.80 ± 1.6e-4 | 5 | 5.00 ± 5.50e-4 |
| outside the array | -3 | -2.98 ± 0.021 | 6 | 5.99 ± 2.00e-3 |
|  | 10 | 10.01 ± 4.00e-5 | 3 | 2.99 ± 6.00e-6 |
|  | 10.8 | 10.82 ± 3.0e-4 | 9.6 | 9.62 ± 2.35e-4 |

https://doi.org/10.1371/journal.pone.0193974.t001

## Acknowledgments

This research was funded by the Environmental Programs Directorate of the Los Alamos National Laboratory. In addition, VVV was supported by the DiaMonD project (An Integrated Multifaceted Approach to Mathematics at the Interfaces of Data, Models, and Decisions, U.S. Department of Energy Office of Science, Grant #11145687. VVV and BSA were supported by LANL LDRD grant 20180060.

## Author Contributions

**Conceptualization:** Valentin G. Stanev, Velimir V. Vesselinov, Boian S. Alexandrov.

**Formal analysis:** Valentin G. Stanev, Velimir V. Vesselinov.

**Methodology:** Boian S. Alexandrov.

**Software:** Filip L. Iliev, Boian S. Alexandrov.

**Validation:** Boian S. Alexandrov.

**Writing ± original draft:** Filip L. Iliev, Velimir V. Vesselinov, Boian S. Alexandrov.

**Writing ± review & editing:** Valentin G. Stanev, Velimir V. Vesselinov, Boian S. Alexandrov.